\pdfoutput=1

\documentclass[11pt]{article}

\usepackage{naacl2021}

\usepackage{times}
\usepackage{latexsym}
\usepackage{multirow}
\usepackage{graphicx}
\usepackage{subfigure}
\usepackage{enumitem}
\usepackage{color, colortbl}
\usepackage{bm}
\usepackage{amsmath}
\usepackage{amsfonts}
\usepackage{array}
\usepackage{url}
\usepackage{amsthm}
\usepackage{amssymb}
\usepackage{xspace}
\usepackage{xcolor, soul}
\sethlcolor{green}
\definecolor{OliveGreen}{rgb}{0,0.6,0}

\newcommand{\tabincell}[2]{\begin{tabular}{@{}#1@{}}#2\end{tabular}}
\newcolumntype{L}[1]{>{\raggedright\arraybackslash}p{#1}}
\newcolumntype{C}[1]{>{\centering\arraybackslash}p{#1}}
\newcolumntype{R}[1]{>{\raggedleft\arraybackslash}p{#1}}
\definecolor{Gray}{gray}{0.9}

\newcommand{\nop}[1]{}

\definecolor{mypink}{rgb}{0.858, 0.188, 0.478}

\newcommand{\model}{\textsc{MQA-QG}\xspace}

\usepackage[T1]{fontenc}

\usepackage[utf8]{inputenc}

\usepackage{microtype}

\title{Unsupervised Multi-hop Question Answering by Question Generation}

\author{Liangming Pan$^{1}$ \quad Wenhu Chen$^{2}$ \quad Wenhan Xiong$^{2}$ \\ \textbf{Min-Yen Kan}$^1$ \quad \textbf{William Yang Wang$^{2}$} \\
$^1$School of Computing, National University of Singapore, Singapore\\
$^2$University of California, Santa Barbara, CA, USA \\
{\tt e0272310@u.nus.edu} \\
{\tt \{wenhuchen, xwhan, william\}@cs.ucsb.edu}\\
{\tt kanmy@comp.nus.edu.sg}\\
}

\begin{document}
\maketitle
\begin{abstract}
Obtaining training data for multi-hop question answering (QA) is time-consuming and resource-intensive. We explore the possibility to train a well-performed multi-hop QA model without referencing any human-labeled multi-hop question-answer pairs, \textit{i.e.}, \textit{unsupervised} multi-hop QA. We propose \model, an unsupervised framework that can generate human-like multi-hop training data from both homogeneous and heterogeneous data sources. \model generates questions by first selecting/generating relevant information from each data source and then integrating the multiple information to form a multi-hop question. Using only generated training data, we can train a competent multi-hop QA which achieves 61\% and 83\% of the supervised learning performance for the HybridQA and the HotpotQA dataset, respectively. We also show that pretraining the QA system with the generated data would greatly reduce the demand for human-annotated training data. Our codes are publicly available at \url{https://github.com/teacherpeterpan/Unsupervised-Multi-hop-QA}. 
\end{abstract}

\section{Introduction}

Extractive Question Answering (EQA) is the task of answering questions by selecting a span from the given context document.  Works on EQA can be divided into the single-hop~\cite{DBLP:conf/emnlp/RajpurkarZLL16,DBLP:conf/acl/RajpurkarJL18,DBLP:journals/tacl/KwiatkowskiPRCP19} and multi-hop cases~\cite{DBLP:conf/emnlp/Yang0ZBCSM18,DBLP:journals/tacl/WelblSR18,DBLP:conf/emnlp/PerezLYCK20}. Unlike single-hop QA, which assumes the question can be answered with a single sentence or document, multi-hop QA requires combining disjoint pieces of evidence to answer a question. 
Though different well-designed neural models~\cite{DBLP:conf/acl/QiuXQZLZY19,DBLP:conf/emnlp/fang20} have achieved near-human performance on the multi-hop QA datasets~\cite{DBLP:journals/tacl/WelblSR18,DBLP:conf/emnlp/Yang0ZBCSM18}, these approaches rely heavily on the availability of large-scale human annotation. Compared with single-hop QA datasets~\cite{DBLP:conf/emnlp/RajpurkarZLL16}, annotating multi-hop QA datasets is significantly more costly and time-consuming because a human worker needs to read multiple data sources in order to propose a reasonable question. 

\begin{figure}[!t]
	\centering
	\includegraphics[width=7.8cm]{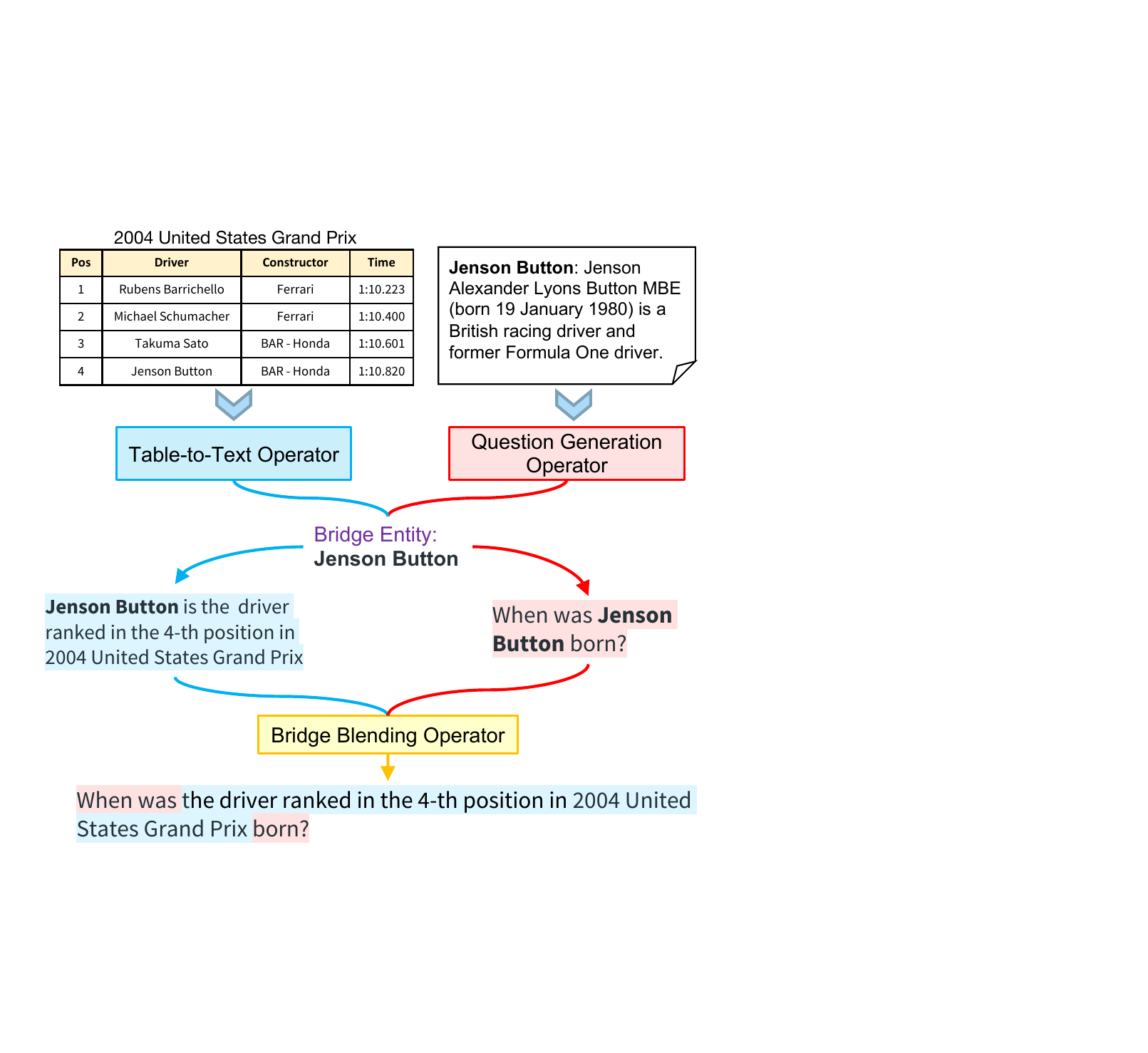}
    \caption{An overview of our approach for generating bridge-type multi-hop questions from table and text. The full set of supported input types and question types are described in Section~\ref{subsec:reasoning_graph}.}. 
    \label{fig:general_framework}
    \vspace{-0.5cm}
\end{figure}


To address the above problem, we pursue a more realistic setting, \textit{i.e.}, \textit{unsupervised} multi-hop QA, in which we assume no human-labeled multi-hop question is available for training, and we explore the possibility of \textit{generating} human-like multi-hop question--answer pairs to train the QA model. We study multi-hop QA for both the homogeneous case where relevant evidence is in the textual forms~\cite{DBLP:conf/emnlp/Yang0ZBCSM18} and the heterogeneous case where evidence is manifest in both tabular and textual forms~\cite{DBLP:conf/emnlp/ChenWH20}. Though successful attempts have been made to generate single-hop question--answer pairs by style transfer~\cite{DBLP:conf/acl/LewisDR19} or linguistic rules~\cite{DBLP:conf/acl/LiWDWX20}, these methods are not directly applicable to the multi-hop setting as: 1) they cannot integrate information from multiple data sources, and 2) they only handle free-form text but not heterogeneous sources as input contexts. 

We propose Multi-Hop Question Generator (\model), a simple yet general framework that decomposes the generation of a multi-hop question into two steps: 1) selecting relevant information from each data source, 2) integrating the multiple information to form a question. Specifically, the model first defines a set of basic \textit{operators} to retrieve / generate relevant information from each input source or to aggregate different information. Afterwards, we define six \textit{reasoning graphs}. Each corresponds to one type of multi-hop question and is formulated as a computation graph built upon the operators. 
We generate multi-hop question--answer pairs by executing the reasoning graph.
Figure~\ref{fig:general_framework} shows an example of generating a \textit{table-to-text question}: a) Given the inputs of (table, text), the $FindBridge$ operator locates a bridge entity that connects the contents between table and text. b) We generate a simple, single-hop question for the bridge entity from the text ($QGwithEnt$ operator) and generate a sentence describing the bridge entity from the table ($DescribeEnt$ operator). c) The $BridgeBlend$ operator blends the two generated contents to obtain the multi-hop question. 

We evaluate our method on two multi-hop QA datasets: HotpotQA~\cite{DBLP:conf/emnlp/Yang0ZBCSM18} and HybridQA~\cite{DBLP:conf/emnlp/ChenWH20}. Questions in HotpotQA reason over multiple texts (homogeneous data), while questions in HybridQA reason over both table and text (heterogeneous data). 
The experiments show that \model can generate high-quality multi-hop questions for both datasets. Without using any human-labeled examples, the generated questions alone can be used to train a surprisingly well QA model, reaching 61\% and 83\% of the F1 score achieved by the fully-supervised setting on the HybridQA and HotpotQA dataset, respectively. We also find that our method can be used in a few-shot learning setting. For example, after pretraining the QA model with our generated data, we can obtain $64.6$ F1 with only $50$ labeled examples in HotpotQA, compared with $21.6$ F1 without the warm-up training. 




In summary, our contributions are: 

\noindent $\bullet$ To the best of our knowledge, this is the first work to investigate unsupervised multi-hop QA. 


\noindent $\bullet$ We propose \model, a novel framework to generate high-quality training data without the need to see any human-annotated multi-hop question. 

\noindent $\bullet$ We show that the generated training data can greatly benefit the multi-hop QA system in both unsupervised and few-shot learning settings. 


\section{Related Work}

\paragraph{Unsupervised Question Answering.} To reduce the reliance on expensive data annotation, \textit{Unsupervised / Zero-Shot QA} has been proposed to train question answering models without any human-labeled training data. 
~\citet{DBLP:conf/acl/LewisDR19} proposed the first unsupervised QA model which generates synthetic (context, question, answer) triples to train the QA model using unsupervised machine translation. However, the generated questions are unlike human-written questions and tend to have a lot of lexical overlaps with the context.  To address this, followup works utilized the Wikipedia cited documents~\cite{DBLP:conf/acl/LiWDWX20}, predefined templates~\cite{DBLP:conf/acl/FabbriNWNX20}, or pretrained language model~\cite{DBLP:conf/emnlp/puri20} to produce more natural questions resembling the human-annotated ones. 


However, all the existing studies are focused on the SQuAD~\cite{DBLP:conf/emnlp/RajpurkarZLL16} dataset to answer \textit{single-hop} and \textit{text-only} questions. These methods do not generalize to multi-hop QA because they lack integrating and reasoning over disjoint pieces of evidence. Furthermore, they are restricted to text-based QA without considering structured or semi-structured data sources such as KB and Table. In contrast, we propose the first framework for unsupervised \textit{multi-hop QA}, which can reason over disjoint structured or unstructured data to answer complex questions. 

\begin{table*}[!t]
    \small
	\begin{center}
	    \renewcommand{\arraystretch}{1.1}
		\begin{tabular}{ l  l  l  l } \hline
        \textbf{Group} & \textbf{Operator} & \textbf{Inputs $\rightarrow$ Outputs} & \textbf{Description} \\ \hline \hline
        \multirow{3}{*}{Selection} & $FindBridge$ & \tabincell{l}{(Table $\mathcal{T}$, Text $\mathcal{D}$) or Texts ($\mathcal{D}_1$, $\mathcal{D}_2$) \\ $\rightarrow$ Bridge Entities $\mathcal{E}^B$} &\tabincell{l}{Select an entity $\mathcal{E}^B$ that links the two input texts \\ $\mathcal{D}_1$ and $\mathcal{D}_2$ (or links the table $\mathcal{T}$ and the text $\mathcal{D}$)} \\ \cline{2-4}
        & $FindComEnt$ & Text $\mathcal{D}$ $\rightarrow$ Comparative Entities $\mathcal{E}^C$ & \tabincell{l}{Extract potential comparative entities from the \\ input text (location, datetime, number, etc.). } \\ \hline
        \multirow{7}{*}{Generation} & $QGwithAns$ & \tabincell{l}{(Text $\mathcal{D}$, Answer $\mathcal{A}$) $\rightarrow$ Question $\mathcal{Q}$} &\tabincell{l}{Generate a single-hop question $\mathcal{Q}$ with answer $\mathcal{A}$ \\ from the input text $\mathcal{D}$} \\ \cline{2-4}
        & $QGwithEnt$ & \tabincell{l}{(Text $\mathcal{D}$, Entity $\mathcal{E}$) $\rightarrow$ Question $\mathcal{Q}$} &\tabincell{l}{Generate a single-hop question $\mathcal{Q}$ that contains \\ the given entity $\mathcal{E}$ from the input text $\mathcal{D}$} \\ \cline{2-4}
        & $DescribeEnt$ & \tabincell{l}{(Table $\mathcal{T}$, Entity $\mathcal{E}$) $\rightarrow$ Sentence $\mathcal{S}$} &\tabincell{l}{Generate a sentence $\mathcal{S}$ that describes the given \\ entity $\mathcal{E}$ based on the information of the table $\mathcal{T}$} \\ \cline{2-4}
        & $QuesToSent$ & \tabincell{l}{Question $\mathcal{Q}$ $\rightarrow$ Sentence $\mathcal{S}$} &\tabincell{l}{Convert a question $\mathcal{Q}$ into its declarative form $\mathcal{S}$} \\ \hline
        \multirow{3}{*}{Fusion} & $BridgeBlend$ & \tabincell{l}{(Question $\mathcal{Q}$, Sentence $\mathcal{S}$, Bridge $\mathcal{E}^B$) \\ $\rightarrow$ Bridge-type multi-hop question $\mathcal{Q}^B$} &\tabincell{l}{Generate a bridge-type multi-hop question $\mathcal{Q}^B$ \\ by fusing the single-hop question $Q$ and the \\ sentence $S$ given the entity $\mathcal{E}^B$ as the bridge} \\ \cline{2-4}
        & $CompBlend$ & \tabincell{l}{(Question $\mathcal{Q}_1$, Question $\mathcal{Q}_2$) $\rightarrow$ \\ Comparative multi-hop question $\mathcal{Q}^C$} &\tabincell{l}{Generate a comparison-type multi-hop question \\ $\mathcal{Q}^C$ by fusing two single-hop questions} \\ \hline
		\end{tabular}
	\end{center}
\caption{The 8 basic operators for \model, categorized into 3 groups. \textbf{Selection}: retrieve relevant information from contexts. \textbf{Generation}: generate information from a single context. \textbf{Fusion}: fuse retrieved/generated information to construct multi-hop questions. Each operator is defined as a function mapping $f(X) \rightarrow Y$.}
\label{tbl:operators}
\end{table*}

\paragraph{Multi-hop Question Generation.}
Question Generation (QG) aims to automatically generate questions from textual inputs~\cite{DBLP:journals/corr/abs-1905-08949}. Early work of Question Generation (QG) relied on syntax rules or templates to transform a piece of given text to questions~\cite{heilman2011automatic,DBLP:conf/coling/ChaliH12a}. With the proliferation of deep learning, QG evolved to use supervised neural models, where most systems were trained to generate questions from \textit{(passage, answer)} pairs in the SQuAD dataset~\cite{DBLP:conf/acl/DuSC17,DBLP:conf/emnlp/ZhaoNDK18,DBLP:conf/aaai/KimLSJ19}. 

With the advent of pretraining language models~\cite{DBLP:conf/nips/00040WWLWGZH19}, the challenge of generating single-hop questions similar to SQuAD  have largely been addressed. QG research has started to generate more complex questions that require deep comprehension and multi-hop reasoning~\cite{DBLP:conf/aaai/Luu20,DBLP:conf/acl/PanXFCK20,DBLP:conf/coling/XiePWKF20,DBLP:conf/www/YuQS020}. For example, \citet{DBLP:conf/aaai/Luu20} proposed a multi-state attention mechanism to mimic the multi-hop reasoning process. \citet{DBLP:conf/acl/PanXFCK20} parsed the input passage as a semantic graph to facilitate the reasoning over different entities. 
However, these supervised methods require large amounts of human-written multi-hop questions as training data. 
Instead, we propose the first \textit{unsupervised} QG system to generate multi-hop questions without the need to access those annotated data. 

\section{Methodology}
\label{sec:method}
The setup of Multi-hop QA is as follows. Given a question $q$ and a set of input contexts $\mathcal{C} = \{ C_1, \cdots, C_n \}$, where each context $C_i$ can be a passage, table, image, etc., the QA model $p_{\theta}(a \vert q, \mathcal{C})$ predicts the answer $a$ for the question $q$ by integrating and reasoning over information from $\mathcal{C}$. 

In this paper, we consider two-hop questions and denote the required contexts as $C_i$ and $C_j$. Formally, each time our model takes as inputs $\langle C_i, C_j \rangle$ to generate a set of $(q, a)$ pairs. We focus on two modalities: the heterogeneous case where $C_i, C_j$ are table and text and the homogeneous case where $C_i, C_j$ are both texts. However, the design of our framework is flexible enough to generalize to multi-hop QA for other modalities. 

Our model \model consists of three components: \textit{operators}, \textit{reasoning graphs}, and \textit{question filtration}. Operators are atomic operations implemented by rules or off-the-shelf pretrained models to retrieve, generate, or fuse relevant information from input contexts $( C_i, C_j )$. Different reasoning graphs define different types of reasoning chains for multi-hop QA with the operators as building blocks. Training $(q, a)$ pairs are generated by executing the reasoning graphs. Question filtration removes irrelevant and unnatural $(q, a)$ pairs to give the final training set $\mathcal{D}$ for multi-hop QA. 

\subsection{Operators}
\label{subsec:basic_operators}

In Table~\ref{tbl:operators}, we define eight basic operators and divide them into three types: 1) \textit{selection}: retrieve relevant information from a single context, 2) \textit{generation}: generate information from a single context, and 3) \textit{fusion}: fuse multiple retrieved/generated information to construct multi-hop questions. 

\paragraph{$\bullet$ \textit{FindBridge}:} Most multi-hop questions rely on the entities that connect different input contexts, \textit{i.e.}, \textit{bridge entities}, to integrate multiple pieces of information~\cite{DBLP:conf/acl-mrqa/XiongYGWCCW19}. \textit{FindBridge} takes two contexts $( C_i, C_j )$ as inputs, and extracts the entities that appear in both $C_i$ and $C_j$ as bridge entities. For example, in Figure~\ref{fig:general_framework}, we extract ``Jenson Button'' as the bridge entity. 

\begin{figure*}[!t]
	\centering
	\includegraphics[width=16cm]{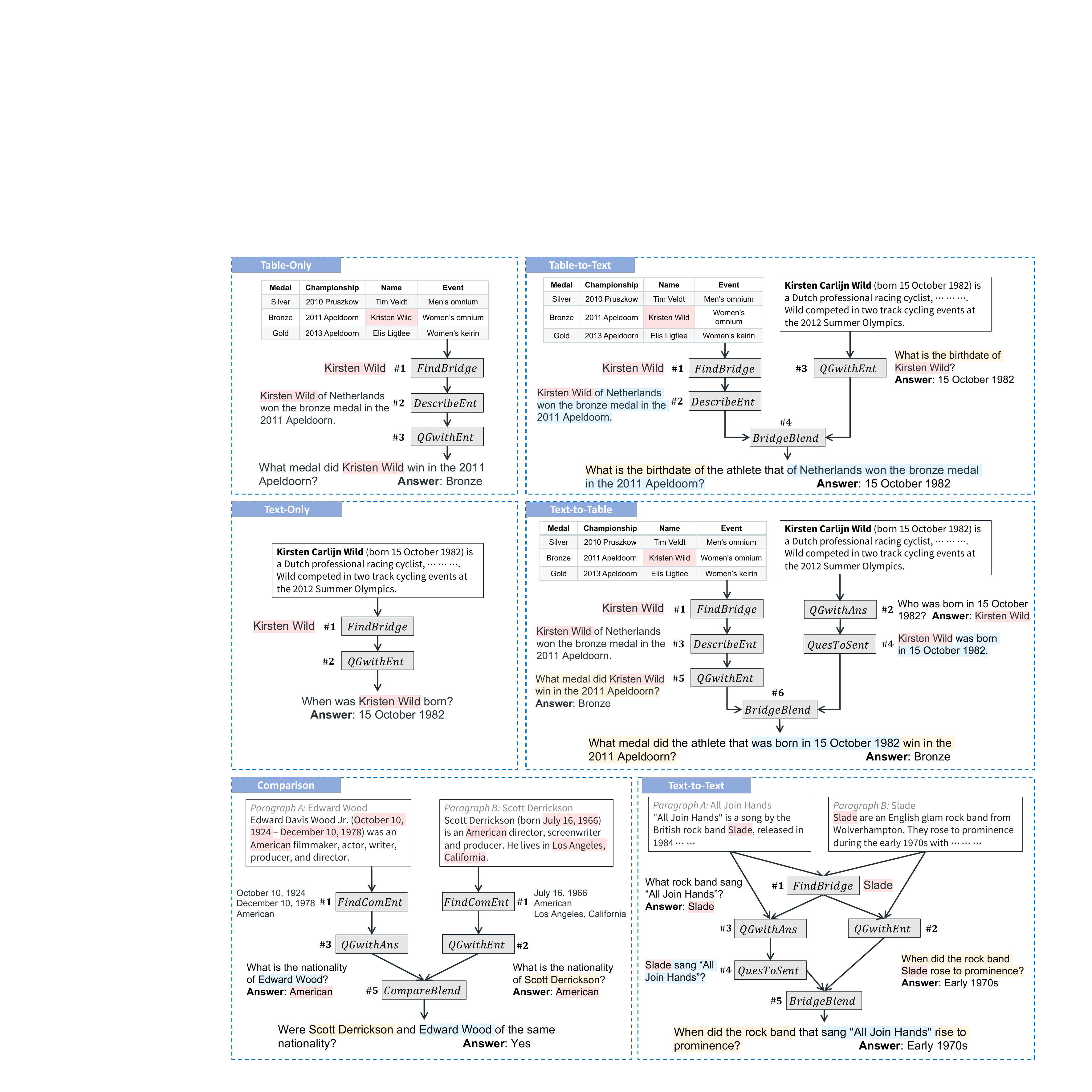}
    \caption{The 6 types of reasoning graphs for \model. Each graph is represented as a DAG of operators. }
    \label{fig:question_blender}
\end{figure*}

\paragraph{$\bullet$ \textit{FindComEnt}:} When generating comparative-type multi-hop questions, we need to decide what property to compare for the bridge entity.  \textit{FindComEnt} extracts potential comparative properties from the input text. We extract entities with NER types $Nationality$, $Location$, $DateTime$, and $Number$ from the input text as comparative properties (\textit{cf}, ``Comparison'' in Figure~\ref{fig:question_blender}). 

\begin{figure}[!t]
	\centering
	\includegraphics[width=7.8cm]{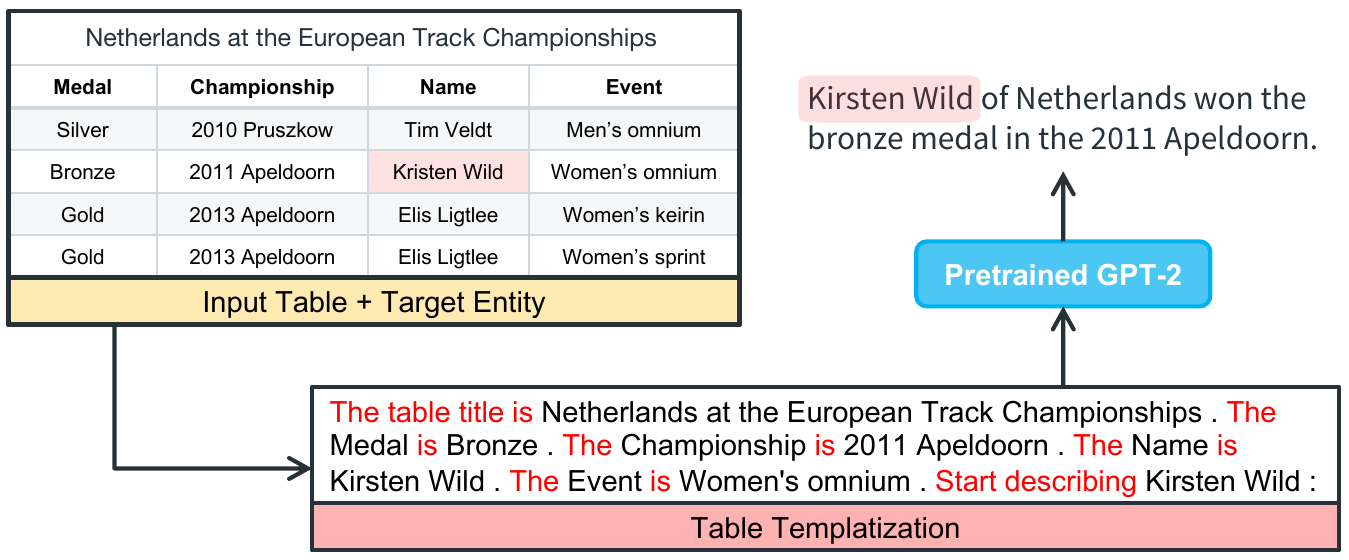}
    \caption{The implementation of \textit{DescribeEnt} operator.}
    \label{fig:table_to_text}
\end{figure}

\paragraph{$\bullet$ \textit{QGwithAns}, \textit{QGwithEnt}:} These two operators generate simple, single-hop questions from a single context, which are subsequently used to compose multi-hop questions. We use the pretrained Google T5 model~\cite{DBLP:journals/corr/abs-1910-10683} fine-tuned on SQuAD to implement these two operators. Given the SQuAD training set of context-question-answer triples $\mathcal{D} = \{(c,q,a)\}$, we jointly fine-tune the model on two tasks. 1) \textit{QGwithAns} aims to generate a question $q$ with $a$ as the answer, given $(c,a)$ as inputs. 2) \textit{QGwithEnt} aims to generate a question $q$ that contains a specific entity $e$, given $(c, e)$ as inputs. The evaluation of this T5-based model can be found in Appendix~\ref{app:evaluate_operators}. 

\paragraph{$\bullet$ \textit{DescribeEnt}:} Given a table $T$ and a target entity $e$ in the table, the \textit{DescribeEnt} operator generates a sentence that describes the entity $e$ based on the information in the table $T$. We implement this using the GPT-TabGen model~\cite{DBLP:conf/acl/ChenCSCW20} shown in Figure~\ref{fig:table_to_text}. The model first uses template to flatten the table $T$ into a document $P_T$ and then feed $P_T$ to the pre-trained GPT-2 model~\cite{radford2019language} to generate the output sentence $Y$. To avoid irrelevant information in $P_T$, we apply a template that only describes the row where the target entity locates. We then finetune the model on the ToTTo dataset~\cite{DBLP:journals/corr/abs-2004-14373}, a large-scale dataset of controlled table-to-text generation, by maximizing the likelihood of $p(Y \vert P_T ; \beta)$, with $\beta$ denoting the model parameters. The implementation details and the model evaluation are in Appendix~\ref{app:evaluate_operators}. 

\paragraph{$\bullet$ \textit{QuesToSent}:} This operator convert a question $q$ into its declarative form $s$ by applying the linguistic rules defined in~\citet{DBLP:journals/corr/abs-1809-02922}. 

\begin{figure}[!t]
	\centering
	\includegraphics[width=7.8cm]{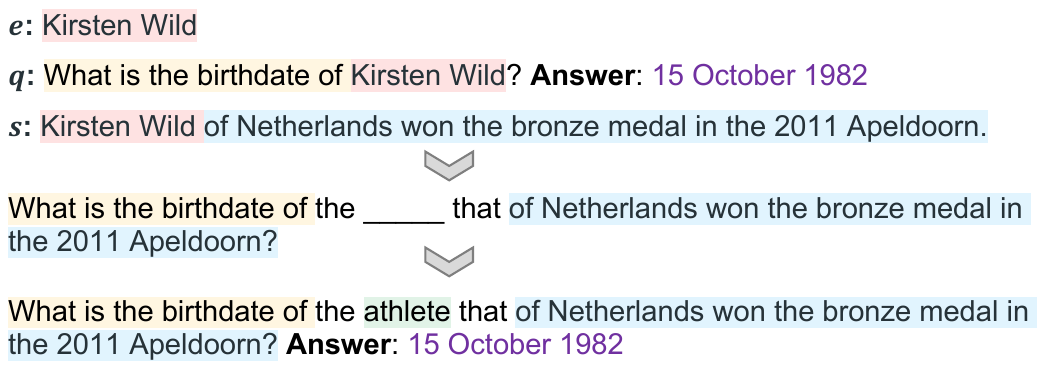}
    \caption{An example of the \textit{BridgeBlend} operator.}
    \label{fig:bridge_blend}
\end{figure}

\paragraph{$\bullet$ \textit{BridgeBlend}:} The operator composes a bridge-type multi-hop question based on: 1) a bridge entity $e$, 2) a single-hop question $q$ that contains $e$, and 3) a sentence $s$ that describes $e$. As exemplified in Figure~\ref{fig:bridge_blend}, we implement this by applying a simple yet effective rule that replaces the bridge entity $e$ in $q$ with ``the [MASK] that $s$'' and employ the pretrained BERT-Large~\cite{DBLP:conf/naacl/DevlinCLT19} to fill in the [MASK] word. 

\paragraph{$\bullet$ \textit{CompBlend}:} This operator composes a comparison-type multi-hop question based on two single-hop questions $q_1$ and $q_2$. The two questions ask about the same comparative property $p$ for two different entities $e_1$ and $e_2$. We form the multi-hop question by filling $p$, $e_1$, and $e_2$ into pre-defined templates (Further details in Appendix~\ref{app:compare_templates}). 

\subsection{Reasoning Graphs}
\label{subsec:reasoning_graph}

Based on the basic operators, we define six types of \textit{reasoning graphs} to generate questions with different types. Each reasoning graph is represented as a directed acyclic graph~(DAG) $\mathcal{G}$, where each node in $\mathcal{G}$ corresponds to an operator. A node $s_i$ is connected by an incoming edge $\langle s_j, s_i \rangle$ if the output of $s_j$ is given as an input to $s_i$. 

As shown in Figure~\ref{fig:question_blender}, \textit{Table-Only} and \textit{Text-Only} represent single-hop questions from table and text, respectively. The remaining reasoning graphs define four types of multi-hop questions. 1) \textit{Table-to-Text}: bridge-type question between table and text, where the answer comes from the text. 2) \textit{Text-to-Table}: bridge-type question between table and text, where the answer comes from the table. 3) \textit{Text-to-Text}: bridge-type question between two texts. 4) \textit{Comparison}: comparison-type question based on two passages. These four reasoning chains can cover a large portion of questions in existing multi-hop QA datasets, such as HotpotQA and HybridQA. We generate QA pairs by executing each reasoning graph. Our framework can easily extend to other modalities and reasoning chains by defining new operators and reasoning graphs. 

\subsection{Question Filtration}
\label{sec:question_filtration}

Finally, we employ two methods to refine the quality of generated QA pairs. 1) \textbf{Filtration}. We use a pretrained GPT-2 model to filter out those questions that are disfluent or unnatural. The top $N$ samples with the lowest perplexity scores are selected as the generated dataset to train the multi-hop QA model. 2) \textbf{Paraphrasing}. We train a question paraphrasing model based on the BART model~\cite{DBLP:conf/acl/LewisLGGMLSZ20} to paraphrase each generated question. Our experiments show that filtration brings noticeable improvements to the QA model. However, we show in Section~\ref{sec:question_paraphrase} that paraphrasing produces more human-like questions but introduces the semantic drift problem that harms the QA performance.

\section{Experiments}
We evaluate our framework on two multi-hop QA datasets: HotpotQA~\cite{DBLP:conf/emnlp/Yang0ZBCSM18} and HybridQA~\cite{DBLP:conf/emnlp/ChenWH20}. HotpotQA focuses on multi-hop QA over homogeneous inputs, while HybridQA deals with multi-hop QA over heterogeneous information. 
HotpotQA contains $\sim$100K crowd-sourced multi-hop questions, where each question requires reasoning over two supporting Wikipedia documents to infer the answer. HybridQA contains $\sim$70K human-labeled multi-hop questions, where each question is aligned with a structured Wikipedia table and multiple passages linked with the entities in the table. The questions are designed to aggregate both tabular information and text information, \textit{i.e.}, lack of either form renders the question unanswerable. 

Table~\ref{tbl:datasets} shows the statistics of these two datasets and Appendix~\ref{app:data_example} gives their data examples. There are two types of multi-hop questions in HotpotQA: bridge-type (81\%) and comparison-type (19\%). For HybridQA, questions are divided by whether their answers come from the table (In-Table question, 56\%) or from the passage (In-Passage question, 44\%). Around 80\% HybridQA questions requires bridge-type reasoning. 


\begin{table}[!t]
  \small
  \begin{center}
      \begin{tabular}{|ccccc|}
      \hline
        Split & Train & Dev & Test & Total \\ \hline \hline
        \multicolumn{5}{|c|}{HotpotQA} \\ \hline
        Bridge & 72,991 & 5,918 & $-$ & 78,909 (81 \%)\\
        Comparison & 17,456 & 1,487 & $-$ & 18,943 (19 \%) \\
        Total & 90,447 & 7,405 & $-$ & 97,852 \\ \hline \hline
        \multicolumn{5}{|c|}{HybridQA} \\ \hline
        In-Passage & 35,215 & 2,025 & 2,045 & 39,285 (56 \%)\\
        In-Table & 26,803 & 1,349 & 1,346 & 29,498 (43 \%) \\
        Compute & 664 & 92 & 72 & 828 (1.1 \%) \\
        Total & 62,682 & 3,466 &  3,463 & 69,611 \\ \hline
      \end{tabular}
  \end{center}
  \caption{Basic statistics of HotpotQA and HybridQA. }
  \label{tbl:datasets}
\end{table}

\begin{table*}[!t]
	\begin{center}
		\begin{tabular}{ c | l | c | c | c } \hline
        \multicolumn{2}{c|}{\multirow{2}{*}{\textbf{Model}}} & \textbf{In-Table} & \textbf{In-Passage} & \textbf{Total} \\ \cline{3-5}
        \multicolumn{2}{c|}{$\ $} & EM / $F_1$ & EM / $F_1$ & EM / $F_1$ \\ \hline \hline
        \multirow{3}{*}{Supervised} 
        & S1. Table-Only~\cite{DBLP:conf/emnlp/ChenWH20} & 14.7 / 19.1 & 2.4 / 4.5 & 8.4 / 7.1 \\
        & S2. Passage-Only~\cite{DBLP:conf/emnlp/ChenWH20} & 9.2 / 13.5 & 26.1 / 32.4 & 19.5 / 25.1 \\
        & S3. \textsc{HYBRIDER}~\cite{DBLP:conf/emnlp/ChenWH20} & \textbf{51.2 / 58.6} & \textbf{39.6 / 46.4} & \textbf{42.9 / 50.0} \\ \hline
        \multirow{3}{*}{Unsupervised} & 
        U1. QDMR-to-Question & 25.7 / 29.7 & 12.8 / 16.5 & 17.7 / 21.4 \\
        & U2. \model \textit{-w/o} Filtration & 33.0 / 37.1 & 18.6 / 23.4 & 23.8 / 28.2 \\ 
        & U3. \model & \textbf{36.2 / 40.6} & \textbf{19.8 / 25.0} & \textbf{25.7 / 30.5} \\ \hline
		\end{tabular}
	\end{center}
\caption{Performance comparison between supervised models and unsupervised models on \textbf{HybridQA}. }
\label{tbl:hybrid_performance}
\end{table*}

\begin{table*}[!t]
	\begin{center}
		\begin{tabular}{ c | l | c | c | c } \hline
        \multicolumn{2}{c|}{\multirow{2}{*}{\textbf{Model}}} & \textbf{Bridge} & \textbf{Comparison} & \textbf{Total} \\ \cline{3-5}
        \multicolumn{2}{c|}{$\ $} & EM / $F_1$ & EM / $F_1$ & EM / $F_1$ \\ \hline \hline
        \multirow{1}{*}{Supervised}
        & S4. SpanBERT~\cite{DBLP:journals/tacl/JoshiCLWZL20} & \textbf{68.2 / 83.5} & \textbf{74.2 / 80.3} & \textbf{69.4 / 82.8} \\ \hline
        \multirow{5}{*}{Unsupervised} & 
        U4. Bridge-Only & 55.4 / 71.4 & 12.4 / 19.1 & 46.7 / 60.9 \\
        & U5. Comparison-Only & 9.8 / 14.5 & 38.2 / 45.0 & 15.5 / 20.6 \\
        & U6. SQuAD-Transfer & 54.6 / 69.7 & 25.3 / 35.2 & 48.7 / 62.8 \\
        & U7. \model \textit{-w/o} Filtration & 55.2 / 71.2 & 44.8 / 52.9 & 53.1 / 67.5 \\
        & U8. \model & \textbf{56.5 / 72.2} & \textbf{48.8 / 54.4} & \textbf{54.9 / 68.6} \\ \hline
		\end{tabular}
	\end{center}
\caption{Performance comparison between supervised models and unsupervised models on \textbf{HotpotQA}. }
\label{tbl:hotpot_performance}
\end{table*}

\subsection{Unsupervised QA Results}
\label{sec:unsupervised_QA}

\paragraph{Question Generation.}
In HybridQA, we extract its table--text corpus consisting of $(T,D)$ input pairs, where $T$ denotes the table and set of its linked passages $D$. We generate two multi-hop QA datasets $\mathcal{Q}_{tbl \rightarrow txt}$ and $\mathcal{Q}_{txt \rightarrow tbl}$ with \model by executing the ``Table-to-Text'' and ``Text-to-Table'' reasoning graphs for each $(T,D)$, resulting in a total of 
170K QA pairs. We then apply question filtration to obtain the training set $\mathcal{Q}_{hybrid}$ with 100K QA pairs. Similarly, for HotpotQA, we first generate $\mathcal{Q}_{bge}$ and $\mathcal{Q}_{com}$, which contains only the bridge-type questions and only the comparison-type questions, respectively. Afterward, we merge them and filter the questions to obtain the final training set $\mathcal{Q}_{hotpot}$ with 100K QA pairs. In Appendix~\ref{app:generated_datasets}, we gives the statistics of all the generated datasets. 

\paragraph{Question Answering} 
For HybridQA, we use the \textsc{HYBRIDER}~\cite{DBLP:conf/emnlp/ChenWH20} as the QA model, which breaks the QA into linking and reasoning to cope with heterogeneous information, achieving the best result in HybridQA. For HotpotQA, we use the SpanBERT~\cite{DBLP:journals/tacl/JoshiCLWZL20} since it achieved promising results on HotpotQA with reproducible codes. We use the standard Exact Match (EM) and $F_1$ metrics to measure the QA performance. 


\paragraph{Baselines.}
We compare \model with both supervised and unsupervised baselines. For HybridQA, we first include the two supervised baselines \textit{Table-Only} and \textit{Passage-Only} in~\citet{DBLP:conf/emnlp/ChenWH20}, which only rely on the tabular information or the textual information to find the answer. As we are the first to target unsupervised QA on HybridQA, there is no existing unsupervised baseline for direct comparison. Therefore, we construct a strong baseline \textit{QDMR-to-Question} that generate questions from Question Decomposition Meaning Representation (QDMR)~\cite{DBLP:journals/tacl/WolfsonGGGGDB20}, a logical representation specially designed for multi-hop questions. We first generate QDMR expressions from the input (table, text) using pre-defined templates and then train a Seq2Seq model~\cite{DBLP:journals/corr/BahdanauCB14} to translate QDMR into question. Details of this baseline are introduced in Appendix~\ref{app:qdmr_to_question}. For HotpotQA, we introduce three unsupervised baselines. \textit{SQuAD-Transfer} trains SpanBERT on SQuAD and then transfers it for multi-hop QA. \textit{Bridge-Only} / \textit{Comparison-Only} use only the bridge-type / comparison-type questions by \model to train the QA model. 

\begin{table*}[!t]
    \small
	\begin{center}
		\begin{tabular}{ c|c|c|c|c|c|c|c|c|c } \hline
        \multirow{3}{*}{Setting} & \multicolumn{4}{c|}{\textbf{Components}} & \multicolumn{2}{c|}{\textbf{Reasoning Types}} & \multicolumn{3}{c}{\textbf{Performance}} \\ \cline{2-10}
        & \multirow{2}{*}{Text} & \multirow{2}{*}{Table} & \multirow{2}{*}{Fusion} & \multirow{2}{*}{Filtration} & \multirow{2}{*}{Table$\rightarrow$Text} & \multirow{2}{*}{Text$\rightarrow$Table} & In-Table & In-Passage & Total \\ \cline{8-10} 
        & & & & & & & EM / $F_1$ & EM / $F_1$ & EM / $F_1$ \\ \hline
        A1 & \checkmark & & & & & & 12.4 / 14.9 & 2.7 / 4.3 & 6.4 / 8.3 \\
        A2 & & \checkmark & & & & & 19.4 / 23.3 & 3.4 / 5.5 & 9.6 / 12.3 \\
        A3 & \checkmark & \checkmark & & & & & 14.8 / 19.2 & 5.6 / 7.8 & 9.1 / 12.1 \\
        A4 & \checkmark & \checkmark & \checkmark & & \checkmark & & 11.1 / 15.2 & 17.3 / 21.9 & 14.9 / 19.4 \\
        A5 & \checkmark & \checkmark & \checkmark & & & \checkmark & 41.5 / 47.9 & 0.2 / 1.9 & 16.2 / 19.8 \\ 
        A6 & \checkmark & \checkmark & \checkmark & & \checkmark & \checkmark & 33.0 / 37.1 & 18.6 / 23.4 & 23.8 / 28.2 \\
        A7 & \checkmark & \checkmark & \checkmark & \checkmark & \checkmark & \checkmark & 36.2 / 40.6 & 19.8 / 25.0 & 25.7 / 30.5 \\ \hline
		\end{tabular}
	\end{center}
\caption{Ablations on the HybridQA development set. \textbf{Text/Table}: whether we utilize the information in the text/table. \textbf{Fusion}: whether we fuse the information from table and text. \textbf{Filtration}: whether we perform question filtration. \textbf{Reasoning Types}: which types of multi-hop questions are generated. }
\label{tbl:ablation_study}
\end{table*}

\paragraph{Performance Comparison.}
Table~\ref{tbl:hybrid_performance} and Table~\ref{tbl:hotpot_performance} summarizes the QA performance on HybridQA and HotpotQA, respectively. For HybridQA, we use the reported performance of \textsc{HYBRIDER} as the supervised benchmark (S3) and apply the same model setting of \textsc{HYBRIDER} to train the unsupervised version, \textit{i.e.}, using our generated QA pairs as the training data (U2 and U3). For HotpotQA, the original paper of SpanBERT only reported the results for the MRQA-2019 shared task~\cite{DBLP:conf/acl-mrqa/FischTJSCC19}, which only includes the bridge-type questions in HotpotQA. Therefore, we retrain the SpanBERT on the full HotpotQA dataset to get the supervised benchmark (S4) and using the same model setting to train the unsupervised versions (U7 and U8). 

Our unsupervised model \model attains 30.5 $F_1$ on the HybridQA test set and 68.6 $F_1$ on the HotpotQA dev set, outperforming all the unsupervised baselines (U1, U4, U5, U6) by large margins. Without using their human-annotated training data, the $F1$ gap to the fully-supervised version is only $19.5$ and $14.2$ for HybridQA and HotpotQA, respectively. In particular, the results of U2 and U3 even outperform the two weak supervised baselines (S1 and S2) in HybridQA. This demonstrates the effectiveness of \model in generating good multi-hop questions for training the QA model. 

\subsection{Ablation Study}
\label{sec:ablation_study}

To understand the impact of different components in \model, we perform an ablation study on the HybridQA development set. In Table~\ref{tbl:ablation_study}, we compare our full model (A7) with six ablation settings by removing certain the model components (A1--A4) or by restricting the reasoning types (A5 and A6). We make three key observations. 

\paragraph{Single-hop questions vs. multi-hop questions.}
A1 to A3 generates single-hop questions using the reasoning graph of \textit{Text-Only} (A1), \textit{Table-Only} (A2), or a union of them (A3). Afterwards, we use them to train the HYBRIDER model and test the multi-hop QA performance. In these cases, the model is trained to answer questions based on either table or text but lacking the ability to reason between table and text. As shown in Table~\ref{tbl:ablation_study}, A1--A3 achieves a low performance of EM and F1, especially for In-Passage questions, showing that single-hop questions alone are insufficient to train a good multi-hop QA system. This reveals that learning to reason between different contexts is essential for multi-hop QA and justifies the necessity of generating multi-hop questions. However, for HotpotQA, we observe that the benefit of multi-hop questions is not as evident as in HybridQA: the SQuAD-Transfer (U6) achieves a relatively good F1 of $62.8$. A potential reason is that the examples of HotpotQA contain reasoning shortcuts through which models can directly locate the answer by word-matching, without the need of multi-hop reasoning, as observed by~\citet{DBLP:conf/acl/JiangB19}. 

\begin{figure*}[!t]
\centering
\subfigure[HybridQA]
{
	\begin{minipage}[t]{0.48\linewidth}
	\centering
	\includegraphics[width=6.8cm]{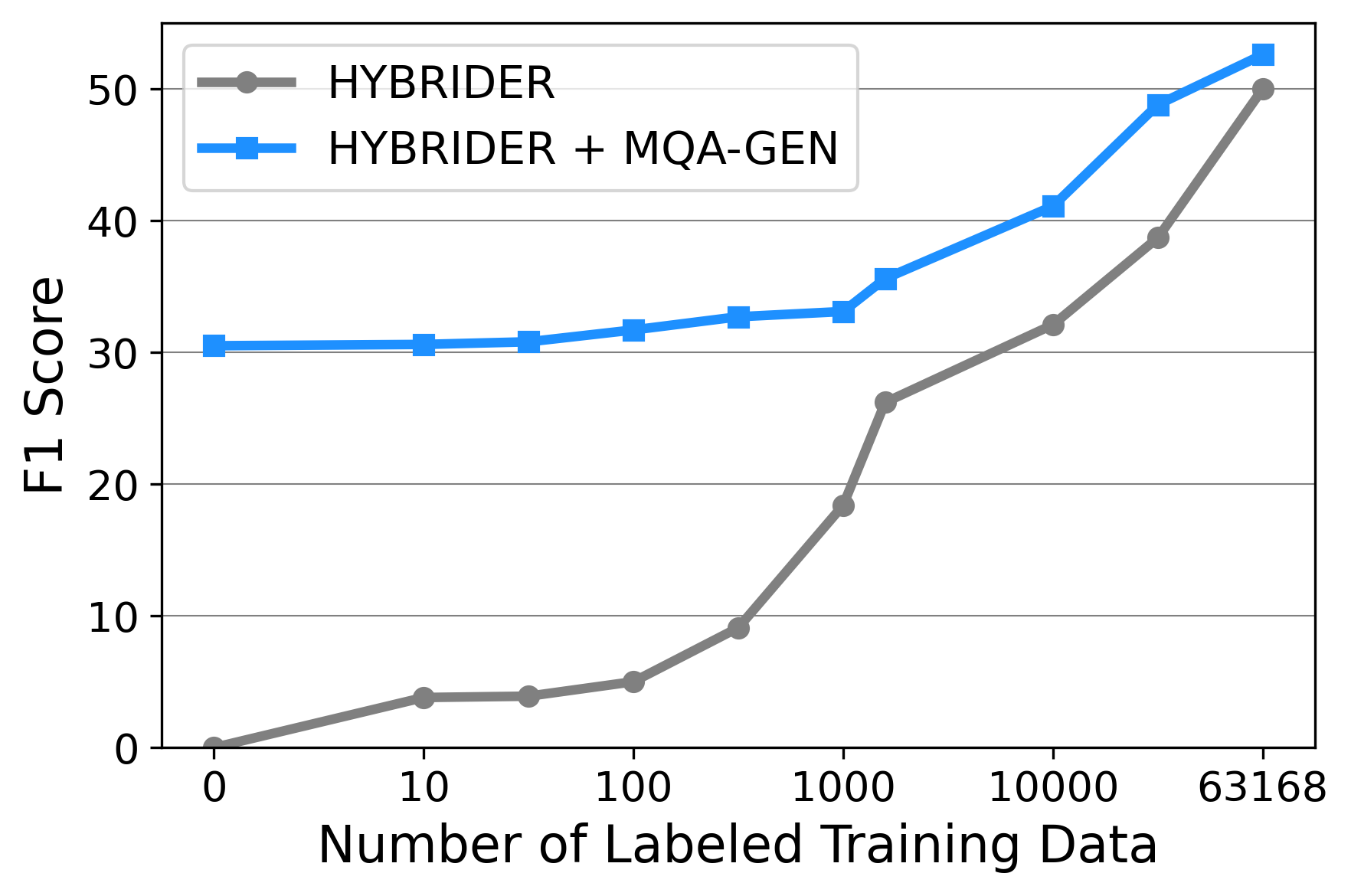}
	\end{minipage}
	\label{fig:few_shot_hybrid}
}
\subfigure[HotpotQA]
{
	\begin{minipage}[t]{0.48\linewidth}
	\centering
	\includegraphics[width=6.8cm]{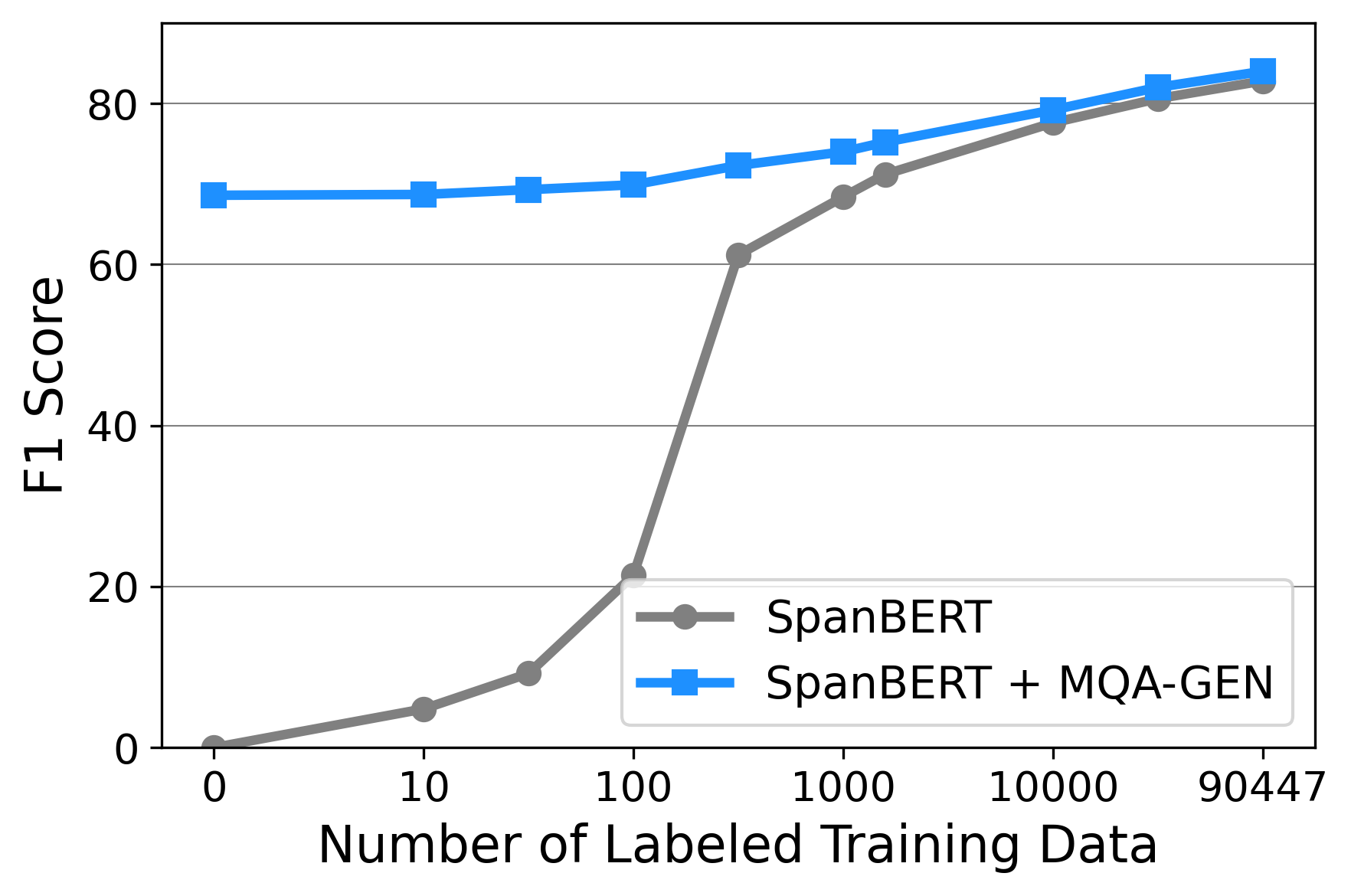}
	\end{minipage}
	\label{fig:few_shot_hotpot}
}
\caption{The few-shot learning experiment. The figure shows the F1 score on the HybridQA (a) / HotpotQA (b) development set for progressively larger training dataset sizes. Note the difference in scales for the Y-axes.}
\label{fig:few_shot}
\end{figure*}

\paragraph{Effect of reasoning types.} 
When we train the model with only the Text-to-Table questions (A5), the model achieves 47.9 F1 for In-Table questions and nearly zero performance for In-Passage questions. However, training with only the Table-to-Text questions (A4) also benefits the In-Table questions (15.2 F1). We believe the reason is that the information in the text can also answer some In-Table questions. Using both reasoning types (A6), the model improves on average by 8.6 F1 compared with the models using a single reasoning type (A4, A5). This shows that it is beneficial to train the multi-hop QA model with diverse reasoning chains. 

\paragraph{Effect of question filtration. }
Question filtration also helps to train a better QA model, leading to a +2.3 F1 for HybridQA and +1.1 F1 for HotpotQA. We find that the GPT-2 based model can filter out most ungrammatical questions but would keep valid yet unnatural questions such as ``Where was the event that is held in 2016 held?''. 

\subsection{Few-shot Multi-hop QA}
\label{sec:fewshot}

We then explore \model's effectiveness in the few-shot learning setting where only a few human-labeled $(q,a)$ pairs are available. We first train the unsupervised QA model based on the training data generated by our best model. Then we fine-tune the model with limited human-labeled data. The blue line in Figure~\ref{fig:few_shot_hybrid} and Figure~\ref{fig:few_shot_hotpot} shows the F1 scores with different numbers of labeled training data for HybridQA and HotpotQA, respectively. We compare this with training the QA model directly on the human-labeled data without unsupervised QA pretraining (grey lines in Figure~\ref{fig:few_shot}). 

With progressively larger training dataset sizes, our model performs consistently better than the model without unsupervised pretraining for both two datasets. The performance improvement is especially prominent in very data-poor regimes; for example, our approach achieves 69.3 F1 with only 100 labeled examples in HotpotQA, compared with 21.4 F1 without unsupervised pretraining (47.9 absolute gain). The results show pretraining QA with \model greatly reduce the demand for human-annotated data. It can be used to provide a ``warm start'' for online learning QA system in which training data are quite limited for a new domain. 

\begin{table*}[!t]
    \small
	\begin{center}
	    \renewcommand{\arraystretch}{1.1}
		\begin{tabular}{ c c l  l } \hline
        \textbf{Type} & \# & \textbf{Generated Question} & \textbf{Answer} \\ \hline
        \multirow{2}{*}{\tabincell{c}{Table-\\to-Text}} & 1 & When did {\color{red} the one} that won the Eurovision Song Contest in 1966 join Gals and Pals? & 1963 \\
        & 2 & How many students attend the teams that played in the Dryden Township Conference? & 1900 \\ \hline
        \multirow{2}{*}{\tabincell{c}{Text-\\to-Table}} & 3 & What album did the Oak Ridge Boys release in 1989? & American Dreams \\
        & 4 & When was {\color{blue} the name that is the name of} the bridge that crosses Youngs Bay completed? & 1921 \\ \hline
        \multirow{2}{*}{\tabincell{c}{Text-\\to-Text}} & 5 & Which Canadian cinematographer is best known for his work on Fargo? & Craig Wrobleski \\
        & 6 & What is illegal in the {\color{red} country} that is Bashar Hafez al - Assad 's father? & Cannabis \\ \hline
        \multirow{2}{*}{Comp.} & 7 & Who was born first, Terry Southern or Neal Town Stephenson? & Terry Southern \\
        & 8 & Are Beth Ditto and Mary Beth Patterson of the same nationality? & Yes \\ \hline
		\end{tabular}
	\end{center}
\caption{Examples of multi-hop question--answers generated by \model, categorized by reasoning graphs. The two major error types are highlighted: red for \textit{inaccurate reference} and blue for \textit{redundancy}. }
\label{tbl:question_examples}
\end{table*}

\begin{figure}[!t]
	\centering
	\includegraphics[width=7.5cm]{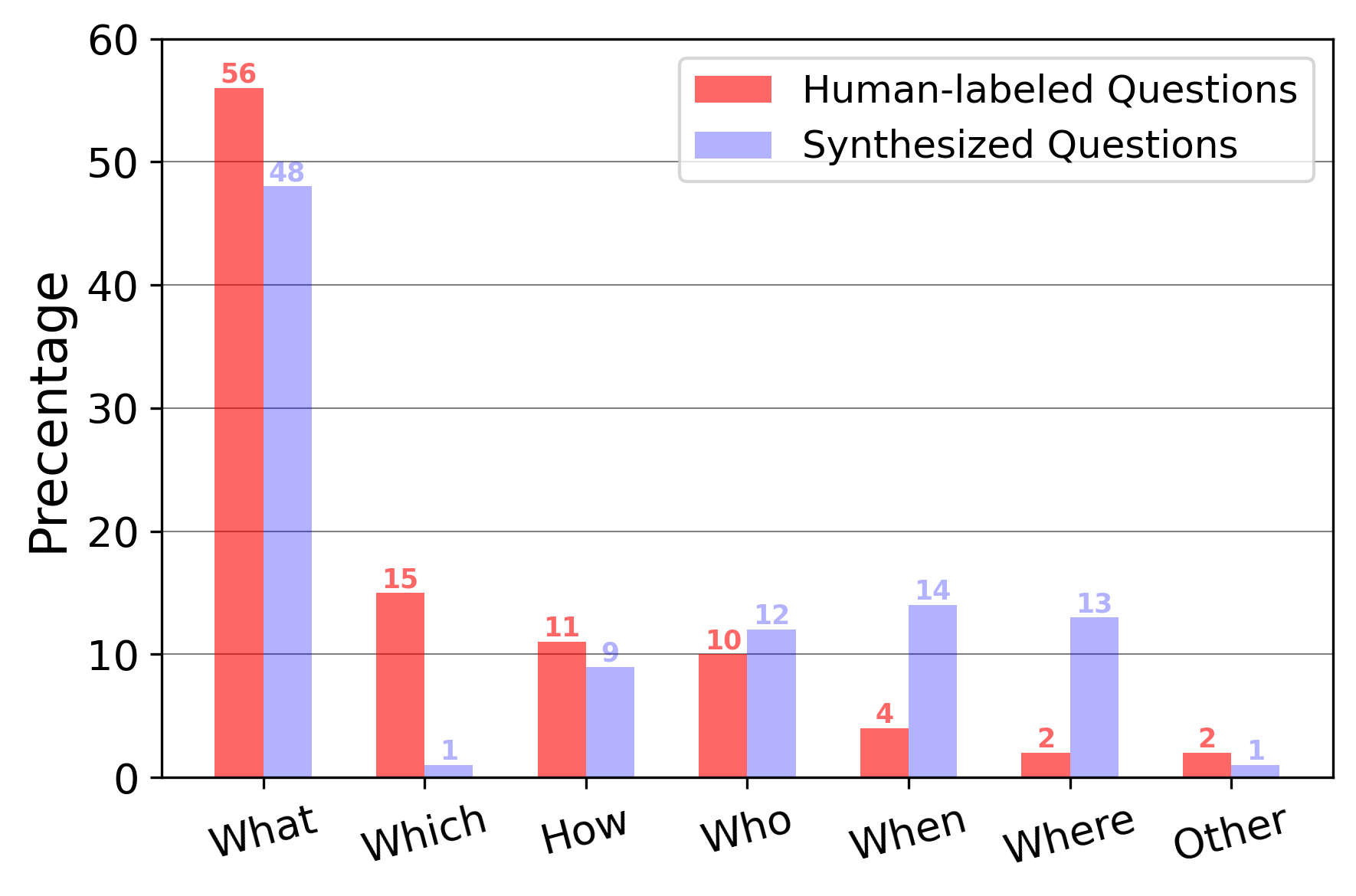}
    \caption{Question type distribution for our generated dataset and the human-labeled dataset for HybridQA.}
    \label{fig:question_types}
\end{figure}

\subsection{Analysis of Generated Questions}
\label{sec:question_analysis}


Although the generated questions are used to optimize for downstream QA performance, it is still instructive to examine the output QA pairs to better understand our system's advantages and limitations. In Figure~\ref{fig:question_types}, we plot the question type distribution for both the human-labeled dataset and the generated data for HybridQA. We find that the two datasets have a similar question type distribution, where ``What'' questions constitute the major type. However, our model generates more ``When'' and ``Where'' questions but fewer ``Which'' questions. This is because the two reasoning graphs we apply for HybridQA are bridge-type questions while ``Which'' questions mostly compare. 

Table~\ref{tbl:question_examples} shows representative examples generated by our model. Most questions are fluent and exhibit encouraging language variety, such as Examples~2, 3, 5. Our model also shows almost no sign of semantic drift, meaning most of the questions are valid despite sometimes being unnatural. The two major deficiencies are \textit{inaccurate references} (in red) and \textit{redundancies} (in blue), shown in Examples~1, 4, 6. This can be addressed by incorporating minimal supervision to guide the fusion process; \textit{i.e.}, more flexible paraphrasing in fusion. 

\begin{table}[!t]
\small
	\begin{center}
		\begin{tabular}{ l | c | c | c } \hline
        \multirow{2}{*}{\centering \textbf{HybridQA}} & \textit{In-Table} & \textit{In-Passage} & \textit{Total} \\ \cline{2-4}
        & EM / $F_1$ & EM / $F_1$ & EM / $F_1$ \\ \hline \hline
        \model & \textbf{36.2 / 40.6} & \textbf{19.8 / 25.0} & \textbf{25.7 / 30.5}  \\ 
        + Paraphrasing & 37.7 / 43.5 & 12.1 / 15.8 & 21.8 / 26.2 \\ \hline \hline
        \multirow{2}{*}{\centering \textbf{HotpotQA}} & \textit{Bridge} & \textit{Comparison} & \textit{Total} \\ \cline{2-4}
        & EM / $F_1$ & EM / $F_1$ & EM / $F_1$ \\ \hline \hline
        \model & \textbf{56.5 / 72.2} & \textbf{48.8 / 54.4} & \textbf{54.9 / 68.6} \\
        + Paraphrasing & 51.7 / 67.0 & 45.7 / 51.1 & 50.5 / 63.8 \\ \hline
		\end{tabular}
	\end{center}
\caption{Unsupervised multi-hop QA performance with/without question paraphrasing. }
\label{tbl:paraphrase_results}
\end{table}

\subsection{Effects of Question Paraphrasing}
\label{sec:question_paraphrase}

As discussed in Section~\ref{sec:question_filtration}, to generated more natural-looking questions, we attempted to train a BART-based question paraphrasing model to paraphrase each generated question. We finetune the pretrained BART model on the Quora Question Paraphrasing dataset\footnote{https://www.quora.com/q/quoradata/First-Quora-Dataset-Release-Question-Pairs}, which contains over 100,000 question pairs with equivalent semantic meaning. 

The evaluation results are shown in Table~\ref{tbl:paraphrase_results}. Surprisingly, we observe a performance drop for both the HybridQA and the HotpotQA dataset, with a $4.3$ and $4.8$ decrease in F1, respectively. We observe that paraphrasing indeed produces more fluent questions by rewriting the redundancy parts of the original questions into more concise expression. However, paraphrasing introduces the ``semantic drift'' problem, \textit{i.e.}, the paraphrased question changes the semantic meaning of the original question. We believe this severally hurts the QA performance because it produces noisy samples with inconsistent question and answer. Therefore, we argue that in unsupervised multi-hop QA, semantic faithfulness is more important than fluency for the generated questions. This explains why we design hand-crafted reasoning graphs to ensure the semantic faithfulness. However, how to generate fluent human-like questions while keeping semantic faithfulness is an important future direction. 

\section{Conclusion and Future Works}
\label{sec:conclusion}

In this work, we study unsupervised multi-hop QA and propose a novel framework \model to generate multi-hop questions via composing reasoning graphs built upon basic operators. The experiments show that our model can generate human-like questions that help to train a well-performing multi-hop QA model in both the unsupervised and the few-shot learning setting. Further work is required to include more flexible paraphrasing at the fusion stage. We can also design more reasoning graphs and operators to generate more complex questions and support more input modalities. 

\section*{Acknowledgments}

This research is supported by the National Research Foundation, Singapore under its International Research Centres in Singapore Funding Initiative. 
UCSB authors are not supported by any of the above projects. 

\bibliography{anthology,custom}
\bibliographystyle{acl_natbib}

\appendix

\clearpage

\begin{table*}[!t]
  \begin{center}
      \begin{tabular}{cl|ccc}
      \hline
        Operator & Model & BLEU-4 & METEOR & ROUGE-L \\ \hline \hline
        \multirow{5}{*}{\tabincell{l}{QGwithAns \& \\ QGwithEnt}}  & NQG++~\cite{DBLP:conf/nlpcc/ZhouYWTBZ17} & 13.51 & 18.18 & 41.60 \\ 
        & S2ga-mp-gsa~\cite{DBLP:conf/emnlp/ZhaoNDK18} & 15.82 & 19.67 & 44.24 \\ 
        & CGC-QG~\cite{DBLP:conf/www/LiuBang20} & 17.55 & 21.24 & 44.53 \\
        & Google-T5~\cite{radford2019language} & 21.32 & \textbf{27.09} & 43.60 \\
        & UniLM~\cite{DBLP:conf/nips/00040WWLWGZH19} & \textbf{23.75} & 25.61 & \textbf{52.04} \\ \hline
        \multirow{3}{*}{DescribeEnt} & Seq2Seq Attention~\cite{DBLP:journals/corr/BahdanauCB14} & 28.31 & 27.61 & 56.63 \\
        & GPT2-TabGen~\cite{DBLP:conf/emnlp/ChenWH20} & 33.92 & 32.46 & 55.61 \\
        & GPT2-Medium~\cite{DBLP:conf/emnlp/ChenWH20} & \textbf{35.94} & \textbf{33.74} & \textbf{57.44} \\ \hline
      \end{tabular}
  \end{center}
  \caption{Performance evaluation of the \textit{QGwithAns}, \textit{QGwithEnt}, and \textit{DescribeEnt} operator for different models. The best performance is in bold. We adopt the Google-T5 and the GPT2-Medium in our model \model. }
  \label{tbl:operator_evaluation}
\end{table*}

\section{Implementation Details of Operators}
\label{app:operators}

In this section, we give the detailed implementation of four key operators, including \textit{QGwithAns}, \textit{QGwithEnt}, \textit{DescribeEnt}, and \textit{CompBlend}. We also separately evaluate their performance. 

\subsection{The \textit{QGwithAns}, \textit{QGwithEnt}, and \textit{DescribeEnt} Operators}
\label{app:evaluate_operators}

In summary, \textit{QGwithAns}, \textit{QGwithEnt} are T5-based question generation model trained on the SQuAD dataset, and \textit{DescribeEnt} is a GPT-2 based model trained on the ToTTo dataset. 

\paragraph{Implementation Details}
For the question generation model (the \textit{QGwithAns} and \textit{QGwithEnt} operators), we use the SQuAD data split from~\citet{DBLP:conf/nlpcc/ZhouYWTBZ17} to fine-tune the Google T5 model~\cite{radford2019language}. We implement this based on the pretrained T5 model provided by \url{https://github.com/patil-suraj/question_generation}. 

For the table-to-text generation model (the \textit{DescribeEnt} operator), we adopt the GPT-TabGen model proposed in~\citet{DBLP:conf/emnlp/ChenWH20}. The model first uses a template to flatten the input table $T$ into a document $P_T$ and then feed $P_T$ to the pre-trained GPT-2 model to generate the output sentence $Y$. We fine-tune the model on the ToTTo dataset~\cite{DBLP:journals/corr/abs-2004-14373}, a large-scale dataset for controlled table-to-text generation. In ToTTo, given a Wikipedia table and a set of highlighted table cells, the objective is to produce a one-sentence description that best describes the highlighted cells. The original dataset contains 120,761 human-labeled training samples and 7,700 testing samples. To implement the \textit{DescribeEnt} operator, we select the ToTTo samples that focuses on describing a given target entity $e$ rather than the entire table, based on the following criteria: 1) the highlighted cells are in the same row and contains the target entity, 2) the description starts with the target entity. This gives us 15,135 training $(T, e, s)$ triples and 1,194 testing triples, where $T$ is the table, $e$ is the target entity, and $s$ is the target description. 

\paragraph{Evaluation Setup}
We employ BLEU-4~\cite{DBLP:conf/acl/PapineniRWZ02}, METEOR~\cite{DBLP:conf/wmt/LavieA07}, and ROUGE-L~\cite{lin2004rouge} to evaluate the performance of our implementation. For question generation, we compare the T5-based model with several state-of-the-art QG models, using their reported performance on the Zhou split of SQuAD. For the table-to-text generation, we compare GPT-TabGen with the Seq2Seq baseline with attention. 

\paragraph{Evaluation Results}
Table~\ref{tbl:operator_evaluation} shows the evaluation results comparing against all baseline methods. For question generation, the Google-T5 model achieves a BLEU-4 of 21.32, outperforming NQG++, S2ga-mp-gsa, and CGC-QG by large margins. This is as expected since these three baselines are based on Seq2Seq and do not apply language model pretraining. Compared with the current state-of-the-art model UniLM, the Google-T5 model achieves comparable results, with slightly lower BLEU-4 but higher METEOR. For the table-to-text generation model, we find that GPT2-TabGen outperforms Seq2Seq with attention by 5.61 in BLEU-4. When switching to GPT-2-Medium as the pretraining model, the BLEU-4 further improves by 2.04. In our final model \model, we use the Google-T5 and the GPT2-Medium in the operators. 

\begin{table*}[!t]
	\begin{center}
	    \renewcommand{\arraystretch}{1.1}
		\begin{tabular}{ c c l c } \hline
        \textbf{Comparative Property} & \# & \textbf{Question Template} & \textbf{Answer} \\ \hline
        \multirow{1}{*}{born, birthdate} & 1 & Who was born first, $e1$ or $e2$? & $e1$ / $e2$ \\ \hline
        \multirow{4}{*}{located, location} & 2 & Are $e1$ and $e2$ located in the same place? & Yes / No \\
        & 3 & Which one is located in $a1$, $e1$ or $e2$? & $e_1$ \\ 
        & 4 & Which one is located in $a_2$, $e1$ or $e2$? & $e_2$ \\
        & 5 & Are both $e1$ and $e2$ located in $a_1$? & Yes / No \\ \hline
        \multirow{4}{*}{nationality, nation, country} & 6 & Are $e1$ and $e2$ of the same nationality? & Yes / No \\
        & 7 & Which person is from $a1$, $e1$ or $e2$? & $e_1$ \\ 
        & 8 & Which person is from $a2$, $e1$ or $e2$? & $e_2$ \\ \hline
        \multirow{4}{*}{live, live place, hometown} & 9 & Are $e1$ and $e2$ living in the same place? & Yes / No \\
        & 10 & Which person lives in $a1$, $e1$ or $e2$? & $e_1$ \\
        & 11 & Which person lives in $a2$, $e1$ or $e2$? & $e_2$ \\ \hline
		\end{tabular}
	\end{center}
\caption{The comparative properties and their corresponding question templates used in the \textit{CompBlend} operator. $a1$ / $a2$ denotes the answer for the single-hop question $Q_1$ / $Q_2$. }
\label{tbl:comparative_templates}
\end{table*}

\subsection{The \textit{CompBlend} Operator}
\label{app:compare_templates}

The inputs of the \textit{CompBlend} operator are two single-hop questions $Q_1$ and $Q_2$ that ask about the same comparative property $p$; for example, $Q_1$ = ``What is the nationality of Edward Wood?'', $Q_2$ = ``What is the nationality of Scott Derrickson'', and $p$ = ``Nationality''. We then identify the entity appearing in $Q_1$ and $Q_2$, denoted as $e_1$ and $e_2$, respectively. To form the multi-hop question, we fill in the comparing entities $e1$ and $e2$ into the corresponding templates that we define for the comparative property $p$. One of the resulting comparison question for the above example is ``Are Edward Wood and Scott Derrickson of the same nationality?''. This paper considers four comparative properties and defined a total number of 11 templates for them, summarized in Table~\ref{tbl:comparative_templates}. 

\section{Dataset Details}

In this section, we give further details for both the HotpotQA and the HybridQA dataset, as well as the generated datasets by our model \model. 

\subsection{HotpotQA and HybridQA Examples}
\label{app:data_example}

\begin{figure*}[!t]
	\centering
	\includegraphics[width=16cm]{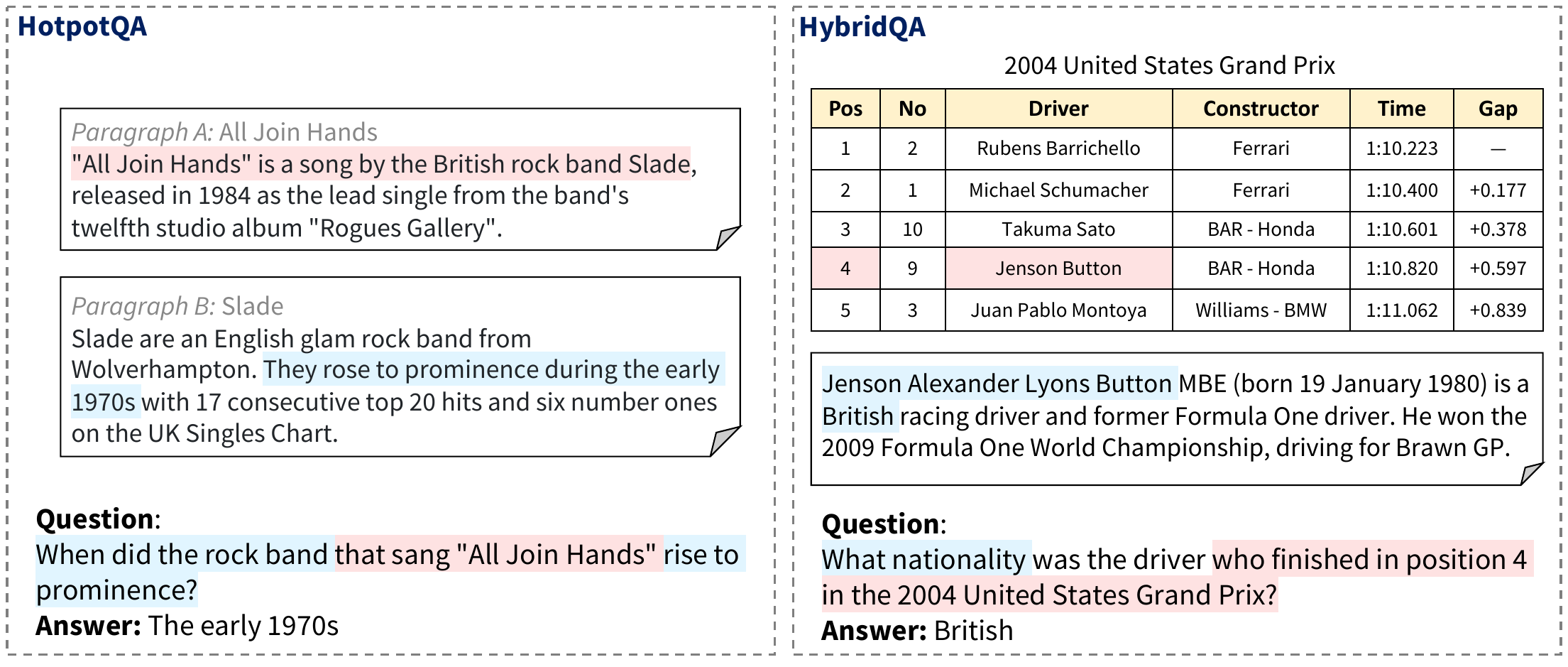}
    \caption{Data examples for the HotpotQA and the HybridQA dataset. Different colors (red and blue) highlight the evidences that are required to answer the multi-hop question from different sources. }
    \label{fig:multihop_example}
\end{figure*}

Figure~\ref{fig:multihop_example} gives data examples for the HotpotQA and the HybridQA dataset. The evidence used to compose the multi-hop question is highlighted, with different colors denoting information from different input contexts. 

\subsection{Statistics of generated datasets}
\label{app:generated_datasets}

For baselines and ablation study, we generate different synthetic training sets by executing different reasoning graphs. For example, we generate two datasets with single-hop questions $\mathcal{Q}_{tbl}$ and $\mathcal{Q}_{txt}$ for HybridQA by executing the ``Table-Only'' and ``Text-Only'' reasoning graphs, respectively. They are applied to train the ablation model A1 and A2. Table~\ref{tbl:generated_datasets} summarizes all the generated datasets generated by our model \model. The column ``Train Model'' denotes each dataset is used to train which model in our experiments. 

\begin{table*}[!t]
  \begin{center}
      \begin{tabular}{|llclc|}
      \hline
        & Dataset & Size & Description & Train Model \\ \hline \hline
        \multirow{4}{*}{HotpotQA} & $\mathcal{Q}_{bge}$ & 129,508 & Bridge-type Questions & U4. Bridge-Only \\
        & $\mathcal{Q}_{com}$ & 115,162 & Comparison-type Questions & U5. Comparison-Only \\
        & $\mathcal{Q}_{bge+com}$ & 244,220 & $\mathcal{Q}_{bge} \cup \mathcal{Q}_{comp}$ & U7. \model \textit{-w/o} Filtration \\ 
        & $\mathcal{Q}_{hotpot}$ & 100,000 & $filtration(\mathcal{Q}_{bge+com})$ & U8. \model \\ \hline
        \multirow{6}{*}{HybridQA} & $\mathcal{Q}_{tbl}$ & 56,448 & Table-Only Questions & A2 \\
        & $\mathcal{Q}_{txt}$ & 47,332 & Text-Only Questions & A1 \\ 
        & $\mathcal{Q}_{txt+tbl}$ & 103,780 & $\mathcal{Q}_{txt} \cup \mathcal{Q}_{tbl}$ & A3 \\
        & $\mathcal{Q}_{txt \rightarrow tbl}$ & 56,448 & Text-to-Table Questions & A5 \\ 
        & $\mathcal{Q}_{tbl \rightarrow txt}$ & 70,661 & Table-to-Text Questions & A4 \\ 
        & $\mathcal{Q}_{txt \leftrightarrow tbl}$ & 127,109 & $\mathcal{Q}_{txt \rightarrow tbl} \cup \mathcal{Q}_{tbl \rightarrow txt}$ & U2. \model \textit{-w/o} Filtration \\ 
        & $\mathcal{Q}_{hybrid}$ & 100,000 & $filtration(\mathcal{Q}_{txt \leftrightarrow tbl})$ & U3. \model \\ \hline
      \end{tabular}
  \end{center}
  \caption{Basic statistics of all the generated datasets by our model \model. }
  \label{tbl:generated_datasets}
\end{table*}

\section{Baseline: QDMR-to-Question}
\label{app:qdmr_to_question}

In this section, we introduce our proposed \textit{QDMR-to-Question}, a strong unsupervised multi-hop QA baseline for HybridQA. We propose this baseline to investigate whether we can generate multi-hop questions from logical forms and compare them with our model \model. 

\paragraph{The QDMR Representation}
The basic idea of QDMR-to-Question is first to generate a structured meaning representation from the source contexts and then convert it into the multi-hop question. We use the Question Decomposition Meaning Representation (QDMR)~\cite{DBLP:journals/tacl/WolfsonGGGGDB20}, a logical representation specially designed for multi-hop questions as the intermediate question representation. QDMR expresses complex questions via atomic operations that can be executed in sequence to answer the original question. Each atomic operation either selects a set of entities, retrieves information about their attributes, or aggregates information over entities. For example, the QDMR for the question ``How many states border Colorado?'' is ``1) Return Colorado; 2) Return border states of \#1; 3) Return the number of \#2''. In contrast to semantic parsing, QDMR operations are expressed through natural language. 

Based on the QDMR representation,~\citet{DBLP:journals/tacl/WolfsonGGGGDB20} crowdsourced BREAK, a large-scale question decomposition dataset consisting of 83,978 (QDMR, question) pairs over ten datasets. 

\paragraph{Multi-hop Question Generation}
Given the table-text $(T,D)$ as inputs, we first generate QDMR representations using two pre-defined templates that represent the \textit{Table-to-Text} question and the \textit{Text-to-Table} question, respectively. The templates with examples are given in Table~\ref{tbl:QDMR_to_question}. We generate QDMRs by randomly filling in the templates. Afterward, we translate the QDMR representation into a natural language question. To this end, we train a Seq2Seq model with attention~\cite{DBLP:journals/corr/BahdanauCB14} on the BREAK dataset, where the input is a QDMR expression, and the target is the corresponding natural language form labeled by humans. We directly apply this Seq2Seq model trained on BREAK as the translator to transform our QDMR representations into multi-hop questions. 

\begin{table*}[!t]
    \small
	\begin{center}
	    \renewcommand{\arraystretch}{1.1}
		\begin{tabular}{ l l l } \hline
        \textbf{QDMR Template} & \textbf{Example} & \textbf{Question} \\ \hline
        \textit{Table-to-Text} & & \multirow{5}{*}{\tabincell{l}{What is the birthdate of \\ the driver that pos is 4 in \\ the 2004 United States \\ Grand Prix?}} \\
        $\quad$ 1) Return $\langle column\ A \rangle$ & 1) Return Driver &  \\
        $\quad$ 2) Return \#1 that $\langle column\ B \rangle$ is $\langle row\ A \rangle$ & 2) Return \#1 in Pos 4 & \\
        $\quad$ 3) Return \#2 in $\langle table \ title \rangle$ & 3) Return \#2 in 2004 United States Grand Prix & \\ 
        $\quad$ 4) Return what is the $\langle text\ attribute \rangle$ of \#3 & 4) Return what is the  birthdate of \#3 & \\ \hline
        \textit{Text-to-Table} & & \multirow{5}{*}{\tabincell{l}{What is the pos of the \\ driver in the 2004 United \\ States that was born in \\ 19 January, 1980? }} \\
        $\quad$ 1) Return $\langle column\ A \rangle$ & 1) Return Driver & \\
        $\quad$ 2) Return \#1 in $\langle table \ title \rangle$ & 2) Return \#1 in 2004 United States Grand Prix & \\
        $\quad$ 3) Return \#2 that $\langle predicate \rangle$ $\langle object \rangle$ & 3) Return \#2 that born 19 January 1980 & \\
        $\quad$ 4) Return what is the $\langle column\ B \rangle$ of \#3 & 4) Return what is the Pos of \#3 & \\ \hline
		\end{tabular}
	\end{center}
\caption{The QDMR templates used in the \textit{QDMR-to-Question} model for HybridQA. }
\label{tbl:QDMR_to_question}
\end{table*}

\begin{figure}[!t]
	\centering
	\includegraphics[width=7.8cm]{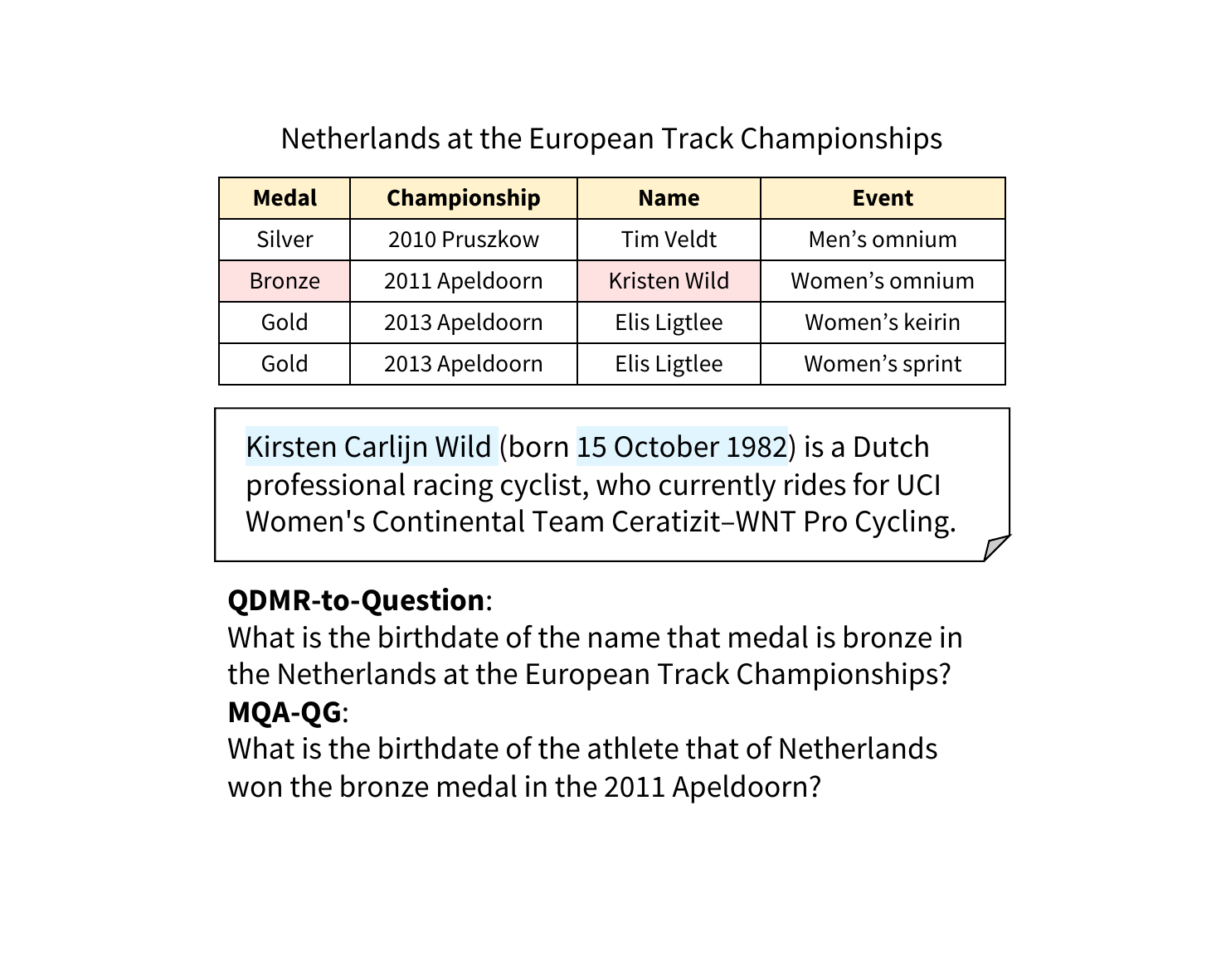}
    \caption{Examples of generated questions for the QDMR-to-Question model and the \model.}
    \label{fig:qdmr_results}
\end{figure}


\paragraph{Evaluation and Discussions}
As shown in Section 4.1, QDMR-to-Question achieves 21.4 F1 on the HybridQA dataset, lower than our model \model by 9.1 F1. A typical example of generated question is shown in Figure~\ref{fig:qdmr_results}. We believe that the main reason for the low performance of QDMR-to-Question is that it lacks a global understanding of the table semantics. Specifically, the model lacks an understanding of the table headers' semantic meaning and the semantic relationship between different headers because table columns and table rows are randomly selected to fill in the QDMR template. For example, in Figure~\ref{fig:qdmr_results}, the model generates an unnatural expression ``the name that medal is bronze'' because it directly copies the table header ``name'' and ``medal'' without understanding them. Instead, as our \model applies the GPT2-based table-to-text model, which encodes the entire table as an embedding, it tends to produce more natural expressions that consider the general table semantics. For the same example, \model generates a better expression ``the athlete that won the bronze medal''. 

\end{document}